\documentclass{article}
\usepackage{graphicx} % Required for inserting images
\usepackage{lipsum}   % for filler text
\usepackage{amsmath}
\usepackage{accents}
\newlength{\dhatheight}
\newcommand{\doublehat}[1]{%
    \settoheight{\dhatheight}{\ensuremath{\widehat{#1}}}%
    \addtolength{\dhatheight}{-0.15ex}%
    \widehat{\vphantom{\rule{1pt}{\dhatheight}}%
    \smash{\widehat{#1}}}}
\usepackage{amsfonts}
\usepackage{url}

\usepackage{natbib}

\title{Leveraging Machine Learning for Official Statistics: A Statistical Manifesto}
\author{Marco J.H. Puts$^1$, David Salgado$^2$, and Piet J.H. Daas$^1$}
\date{
$^1$Department of Methodology, Statistics Netherlands, CBS-weg 11, 6412 EX, Heerlen, The Netherlands\\%
$^2$Department of Methodology, Statistics Spain, Avenida de Manoteras 50-52, 28050 Madrid, Spain \\[2ex]%
February 2024}

\begin{document}

\maketitle

\section{Introduction}
\subsection{Production of official statistics and machine learning}
Evaluating the impact and utilization of machine learning (ML) in the production of official statistics presents an ongoing challenge. ML is a subfield of Artificial Intelligence, which aims at \textquotedblleft not just to understand but also to \emph{build} intelligent entities\textquotedblright\ \citep{RusNor10a}. It is thus similar to assessing the impact of intelligent human activity on a production system, which is limitless. ML itself is composed of different subfields about how the process of learning is carried out: supervised learning, unsupervised learning, reinforcement learning, etc. \citep{IntroMLbook, Mur13a}.

Despite the widespread adoption of ML, implementation still has many challenges \citep[discusses this subject]{Neil2016}. Despite the danger and complexity of ML, the compelling 'datafication' of our society forces us to look at ML as an addition to our (official) statistical toolbox. Datasets get larger, are more detailed, and become more and more complex. Because of this, it becomes increasingly difficult to perform (classical) statistical analysis on these kinds of data. The need for algorithmic-based approaches, which can handle larger, more complex, and unstructured datasets, is necessary to be able to perform successful analysis; see \cite{breiman_statistical_2001}. Without it, we would not be able to extract statistical information from many big data sources, like web scraped data \citep{DaasDoef2020} and aerial images \citep{dejong2020}. 

%Machine learning consists of many techniques. Within the field of data science, these techniques are often called methodology. We don´t agree with the way by which this term is used. Within social sciences and econometrics, the meaning of the terms “methodology”, “methods” and “techniques” are clearly defined and it would be good to stick to these definitions.
ML encompasses various techniques, often referred to as 'a methodology' within the realm of data science. However, most statisticians, as well as most scientists, would disagree with the usage of this term. In social sciences and econometrics, "methodology," "methods," and "techniques" carry specific and distinct meanings, which we advocate for adhering to, at least when talking about ML from an (official) statistical point of view.

When we look at the term 'methods' this becomes most apparent. We have observed that the term 'methods' is incorrectly used in many fields that apply ML. It is often used to merely describe the chosen environment in which a study is performed. So, when summarizing the ML algorithms and hyperparameters used, maybe with some kind of rationale, many data scientists assume this describes the 'method' used. Such a description, however, falls short when viewed from the statistical standpoint of a methodologist. 

So let's start with the basis. Techniques, and how we combine them, are primarily determined by the 'why' and 'what' questions of the application, a.o.: 
\begin{itemize}
    \item why are we doing it? 
    \item why do we choose certain techniques? 
    \item what are we going to do? 
    \item what is our ground material? 
    \item what is the context? 
\end{itemize}

It is the 'how' question that is at the core of the techniques themselves: how do we go about performing certain steps? In addition to the algorithmic description, it describes the (pre- and post-)conditions for applying the technique.
From this, we can define a method as:
\begin{quote}
A method is a systematic procedure of techniques for accomplishing a certain goal. Most of the time these methods are established.
\end{quote}
and a technique as:
\begin{quote}
A technique is a way of carrying out a task. Most of the time these techniques are described as algorithms.
\end{quote}
For example, preparing a meal involves cutting vegetables, boiling eggs, and grilling steaks. Recipes can be considered methods. it is also possible to consider a method that takes into account the context in which one prepares a meal, as well as the circumstances that may arise (for example, a guest may be vegan or allergic to certain ingredients) when preparing a meal. What is the best way to use a specific technique with a specific set of ingredients under what circumstances? Methodology is the subject of this area. When it comes to ML, this implies that we must pay greater attention to the context of methodology when discussing it in relation to its application.\\

Admittedly, sometimes it is hard to determine if something is a method or a technique. Since methods are procedural, they might be interpreted as an algorithm and the other way around. It’s therefore easy to understand why there is confusion about methods and techniques. By definition, they are so closely related that it is easy to confuse them, but for reasons that will become clearer through the rest of this chapter, it is crucial to keep them separate.\\

A priori every production task is susceptible to being impinged by the growing success of ML and deep learning techniques. At least we distinguish two broad groups of production activities potentially affected by these techniques. On the one hand, we have the inference problem providing us with a set of statistics and indicators describing some fragment of social or economic reality. This is, in our view, the core of the business of official statistical production. On the other hand, complementarily important, we have numerous additional tasks improving the quality of the first goal such as data collection, coding, statistical dissemination, etc. In both cases, by and large, the use of ML boils down to building a predictive model to be applied to new data, thus constituting a process step in the whole production cycle \citep[See chapters 23-35 in \cite{Snijkersetal2023}:][]{Snijkers_Dumpert2023,Snijkers_Measure2023,Snijkers_moscardi2023}. For example, predicting values of a continuous target variable will be useful in building model-assisted estimators (as a concrete illustrative example of a regression task in the first group). Also, an automatic coding machine will provide a predicted category for a given statistical unit (as a concrete illustrative example of a classification task in the second group).

In this chapter, we shall focus on supervised learning so that different algorithms will be trained, validated, and tested on a given dataset to be then applied to new data \citep[see e.g.][]{HasTibFri09a}.

\subsection{Production of official statistics and quality}
Quality has been the spinal cord of the production of official statistics making it possible to be used for highly relevant policy-making actions in all countries and the international community \citep[see e.g.][and references therein]{UNNQAF19a}. Nowadays, quality is a multidimensional concept \citep{KARR2006137} and strongly oriented towards users' needs and purpose \citep{eurostat2022}. Measuring and determining quality is thus very important. As a consequence, there exist complementary frameworks developed to assess different aspects of the quality of the production of official statistics \citep[see e.g.][and references therein]{Gootzen2023}.\\

We may cite the output quality approach of many statistical systems \cite[see e.g.][in the context of the European Statistical System]{eurostat2022}, which focuses on the assurance of multiple quality dimensions for the statistical outputs. In the light of the use of ML techniques, \citet{Puts2021} describe how their usage can influence the quality of official statistics in several quality dimensions, namely relevance, accessibility and clarity, coherence and comparability, and accuracy and reliability \citep[see][for definitions, specifically principles 11, 12, 14, and 15]{eurostat2022}. \citet{Puts2021} argue that the following topics must be worked on regarding the methodology of ML in official statistics:
\begin{itemize}
    \item Accuracy and reliability 
    \begin{itemize}
        \item Methodology concerning the human annotation of data
        \item Sampling the population to obtain representative training sets
        \item Using stratification in the context of ML
        \item Correcting the bias caused by the ML model
        \end{itemize}
    \item Accessibility and clarity
    \begin{itemize}
        \item Data structure engineering and selection to increase the transparency of models
        \item Explainability of ML models (explainable AI)
    \end{itemize}
    \item Coherence and comparability
    \begin{itemize}
        \item Reducing spurious correlations
        \item Methodology for studying causation
        \item Dealing with concept drift (representativity over time)
    \end{itemize}
\end{itemize}

Taking the European Statistics Code of Practice \citep{eurostat2022} as a starting point, \citet{saidani2023qualitatsdimensionen} provide an overview of all the attempts made to create an ML quality framework for official statistics. Similar to \cite{Puts2021}, they also conclude that the ESCoP principles 11-15 (statistical output) are crucial in defining the quality of ML algorithms, with the additional consideration of principles 7-10 (statistical processes). They add to this an extra quality dimension, called robustness. As producers of official statistics, we need to make sure that the models are robust for changes in the population.\\

However, in the following, we shall focus on the Total Survey Error Model (TSEM henceforth) \citep{groves2010total} as the starting point for our proposal to adapt this framework to the use of ML in the production of official statistics. As stated above, we shall simplify the use of these techniques to the construction of a predictive model either of continuous, semi-continuous, or categorical variables trained, validated, and tested on a given dataset to be applied to new data for any purpose (thus covering both the core and additional groups of tasks).\\

The TSEM is a comprehensive framework used in survey research and practice to understand and quantify the various sources of error that can affect the accuracy of survey estimates (thus we are focusing on the reliability quality dimension). The approach acknowledges that no survey is perfect, and errors can arise at different stages of the survey process. The goal of the model is to identify, measure, and minimize these errors to improve the overall quality of survey data. This framework implicitly assumes that design-based inference is used to construct estimators and their accuracy assessment (confidence intervals, variance estimation, \dots).\\

\begin{figure}[htbp]
  \centering
  \includegraphics[width=1\textwidth]{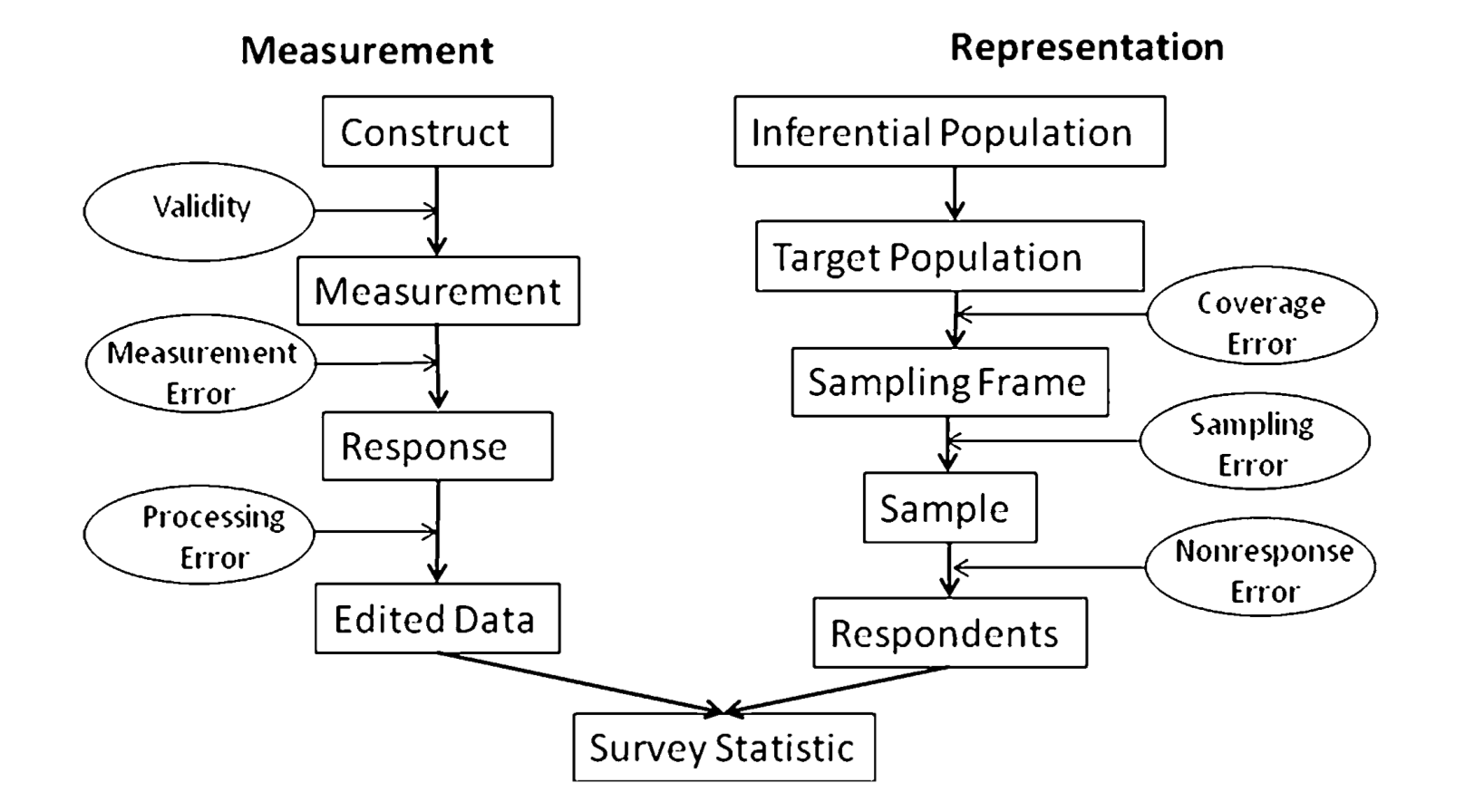} % Adjust the width as needed
  \caption{The total survey error model (source: \citealt{groves2010total})}
  \label{fig:groves}
\end{figure}

In the TSEM, the concepts of population unit and target variable are central, depicted by the so-called representation line and measurement line (see figure \ref{fig:groves}). As such, the model identifies measurement-related errors and representation-related errors. The measurement-related errors are associated to what is being measured, whereas the representation-related errors are about the population and its units and how they are included in the sample. 

We give a brief overview of the definitions of the different sources of error in the TSEM:
The measurement-related errors are:
\begin{itemize}
\item (Construct) Validity error: \\Construct validity errors are related to whether a survey instrument is effectively measuring the theoretical construct or concept it is intended to measure. It assesses the degree to which the survey questions align with the underlying construct of interest.\\
It is caused by an inadequate alignment between survey questions and the conceptual framework being studied can lead to construct validity concerns.
\item Measurement error:\\
Measurement errors occur when there is a discrepancy between the true value of a variable and the value obtained through measurement. This can result from respondent misunderstanding, misreporting, or errors in the survey instrument.\\
It is caused by respondent biases, unclear survey questions, or issues with data collection instruments.
\item Processing error:\\
Processing errors involve mistakes made during the data collection and data processing stages, including errors in data entry, coding, and editing.\\
It is caused by human error in the handling and processing of survey data.
    
\end{itemize}

The representation-related errors are:
\begin{itemize}
\item Coverage error:\\
Coverage errors arise when the sampling frame does not accurately represent the target population. It includes individuals who should be in the population but are not in the frame (undercoverage) or individuals in the frame who are not in the population (overcoverage).
It is caused by an incomplete or inaccurate sampling frame.
\item Sampling Error:\\
Sampling errors occur due to the inherent variability that arises when a sample is used to estimate the characteristics of a population. It is the difference between a sample statistic and the true population parameter. It is caused by random chance in the selection of the sample.
\item Nonresponse Error:\\
Nonresponse errors result from differences between respondents and nonrespondents. It occurs when individuals chosen for the survey do not participate, and their characteristics differ from those who do participate.
It is caused by refusals, inability to reach respondents, or other factors leading to nonparticipation.

\end{itemize}

\citet{Puts2022} argued that such a model can also be valid for ML. First of all, the training set is a sample of the population, the measurements are imperfect (e.g.\ the result of annotations) and can contain errors, and the parameters of a trained model can be seen as estimates. They argued that, due to the presence of both measurement errors and representation errors in the training dataset with respect to the population, the model will generate errors and most of these errors will result in biases when applied to new data.\\

In contrast, \cite{saidani2023qualitatsdimensionen} claim translation of TSEM errors into ML presents a challenge. This observation is fundamentally accurate. TSEM was originally designed for evaluating survey methodology errors, but adapting it to ML presents a number of challenges. Moreover, quantifying the various errors outlined in the TSEM is difficult, and achieving this task would, of course,  exceed the scope of this chapter.

In this document, we propose including a TSEM-like view on the construction of a supervised learning model as a means to assess the quality of the model. It will be referred to as the Total Machine Learning Error model. While the framework will be completed to the greatest extent possible for the time being, there might be some gaps that will need to be filled, thus providing further opportunities for research and refinement in the future. 

\subsection{Outline of the chapter}
The remainder of this chapter is structured as follows. After the introduction, the next section focuses on populations and samples. Here, the viewpoint of Deming on this topic provides valuable insights. This is followed by a description of the Total Machine Learning Error (TMLE) Model and the various phases discerned. In section 4 an overview is given of the consequences of the TMLE-model which provides valuable insights. This section is followed by an overview of some ML applications that both inspired and benefited from the development of the TMLE-model. The chapter ends with a discussion in which the most important topics for future research are listed.\\

\section{Deming´s machine: populations and samples}
\label{sec:DemMachine}
For our purposes, let's start with an example used by \citet{Dem42a} (later on suggested as an exercise in his well-known textbook about survey sampling \citep{Dem50a}). An industrial business person owns an industrial machine producing bolts with a set of technical specifications (weight, size, lengths, resistance to temperature, etc.). Every, say, $N=100$ units are packed up in a box to be sold to retailers. For evident reasons, this person can pose two complementary and different questions in this situation: (a) how many defective bolts (failing technical specifications) exist in each box of $N=100$ bolts? and (b) how many defective bolts are produced on average by the machine? Notice that these are a priori independent concerns rightfully relevant in this situation. This simple example will allow us to introduce relevant statistical and mathematical concepts which will be the basis for our proposed TMLE-model for the production of official statistics.\\
	
Firstly, notice that in question (a) no random element exists in the formulation of the concern: for each box, say $U_{i}$, with a specific known number of bolts, say $N_{i}$, there exists a fixed, but unknown number of defective bolts, say $N_{i}(D)\leq N_{i}$. All we want is to know $N_{i}(D)$ to monitor the output quality of our production system. There is no reference to any probability distribution underlying the box and quantities such as total, mean, and variance must be understood as numerical aggregation figures much in the line of exploratory data analysis. The characteristics of each bolt, i.e.\ the target variables\footnote{In our simplified case, the defectiveness binary indicator $\delta_{k}(D)\in\{0, 1\}$.} are fixed but unknown numbers.\\
	
However, question (b) contains an implicit reference to an underlying random process or random experiment every time a bolt is produced by the machine. The question is meaningful only when randomness is recognized to be present in the generation process of each bolt so that the result may be different, i.e.\  sometimes defective sometimes working. The interest is not only focused on the generation mechanism in the past but genuinely on (immediately) future instances of the generation process.\\ 
	
This distinction allows us to formalize and motivate the following definitions, which are implicitly used at all times in similar related statistical analyses. In the context of question (a) a finite population $U$ is a set of identifiable units $u_{k}$ which we usually denote by their labels $k$ so that $U=\{1,\dots, N\}$. In this line, a sample $s$ is just a subset of units selected from $U$, i.e. $s\subset N$ \citep[see e.g.][]{CasSarWre77a}. Ordered and/or with-replacement samples \citep{Koo74a} can be further expressed in rigorous mathematical language but the key underlying concept is the same: a finite population is a collection of population units; no probability measure is involved in the definition; it is a set-theoretic concept. The characteristics of interest $\mathbf{y}_{k}$ of these population units are just numeric variables, i.e.\ $y_{kq}\in\mathcal{D}^{(q)}$ for each variable $q=1,\dots, Q$, where $\mathcal{D}^{(q)}$ stands for the numerical set of possible values of variable $y^{(q)}$.\\
	
In the context of question (b), more subtleties are needed. The underlying randomness (and probability space) enters into play by defining the (infinite) population as the probability distribution function $F_{{\boldsymbol \theta}}$ of variables $\mathbf{y}$, thus now turned into random variables $\mathbf{Y}$. This distribution function $F_{{\boldsymbol \theta}}$ basically concentrates the random generation mechanism of the values $\mathbf{y}_{k}$ for each bolt $k$, in particular, for the defectiveness indicator variable $\delta(D)$. The concept of population unit in this context is much subtler since the mathematical definition of a random variable does not make any reference to such a concept. It is a modelling assumption. In our simplified modelling scenario for the machine, we may conceive of each variable $\delta_{k}(D)$ as a realization of the same binary random variable $\delta(D)\simeq F_{\boldsymbol{\theta}}$. In this way, every time the random experiment is conducted (generation of a bolt), the random variable is realized (as when we toss a coin) and a new value $\delta_{k}(D)$ is generated. This motivates the following definition of a sample \citep{CasBer02a}: \textquotedblleft the random variables $Y_{1},\dots, Y_{n}$ are called a \emph{random sample of size $n$} from the population $F_{\boldsymbol{\theta}}$ if $Y_{1},\dots,Y_{n}$ are mutually independent random variables and the marginal distribution function of each variable $Y_{k}$ is the same function $F_{\boldsymbol{\theta}}$\textquotedblright. Alternatively, they are called independent and identically distributed random variables. Following these definitions, a population unit in the context of question (b) amounts to each instance the random experiment is conducted giving rise to a new bolt, thus identified with this bolt. Notice how the concept of infinite population naturally fits in this description: we have, on the one hand, already generated bolts but, on the other hand, we may also generate as many as we want since they come from a random experiment.\\

In sum, the difference between questions (a) and (b) stems from the modelling assumption of conceiving target variables values $\mathbf{y}_{k}$ as realizations of random variables or not, i.e.\ whether there exists an underlying random generation mechanism or not. In this line of reasoning, we can observe the two fundamental concepts to assess quality in the production of statistics using the TSEM, namely, population and variable. Assessing the quality of both concepts amounts to assessing the quality of the final statistics. This is what the TSEM does for the production of official statistics in the context of question (a), thus motivating the design-based inference paradigm \citep[see e.g.][and multiple references therein]{Til20a}, where no assumption for an underlying random generation mechanism is made for the target variables.\\

This analysis does not mean that more complex modelling assumptions cannot be made or alternative inferential paradigms cannot be followed \citep{ChaCla12a, Lit12a}. Our proposal concentrates on how to fit the increasing use of ML models into the classical quality assessment framework provided by the TSEM.

\section{The total machine learning error model}

To propose a TMLE-model, we shall assume that we are still providing answers to question (a) in Deming's machine scenario, which we understand as the traditional realm of the production of official statistics (typically by statistical offices). This in contrast to question (b), which we understand as the natural realm of the analysis of official statistics (typically by policy-makers, analysts, researchers, and stakeholders in general). This is our reading of the difference between enumerative vs.\ analytic surveys by \citet{Deming41} and \citet{Deming42, Deming53}.\\

The challenge we face is two-fold. Firstly, we intend to provide a proposed TMLE-model for supervising statistical learning models independently of their specific use in a business function in the production process \citep[see][in this book for some examples]{BarSaeSalSan24a}. Secondly, we propose a combination with the TSEM to produce statistics in a general fashion.\\ 

Like the TSEM, the TMLE-model deals with the total error on an ML model as a result of different errors introduced during the process of creating the training set and assessing the model with the test set to be subsequently applied to our target population to produce the statistics of interest. In the TSEM the basic assumption is the existence of true values for the target variables \citep{groves2010total}. Likewise, for statistical learning models we shall assume the existence of a true statistical model expressing the random generation mechanism of a target variable $Y$ from auxiliary variables $\mathbf{X}$, so that with little loss of generality, we may write $Y = f(\mathbf{X}; \boldsymbol{\Theta}) + \epsilon$, where $\boldsymbol{\Theta}$ represent any parametrization of the functional dependence $f$. The model output will basically be an estimated function $\hat{f}$ with a estimated set of parameters $\widehat{{\boldsymbol{\Theta}}}$ so that $\widehat{Y}=\hat{f}\left(\mathbf{X};\widehat{{\boldsymbol{\Theta}}}\right)$. 
The choice of $\hat f$, 
and the resulting parameters 
$\widehat{{\boldsymbol{\Theta}}} = \boldsymbol{\Theta} + \epsilon_\Theta$, 
will lead to errors in the final predictions of the model. Given the model $\hat f$, it is in our opinion essential to identify the error $\epsilon_{\boldsymbol{\Theta}}$, since it will deliver a substantive contribution to the total error of the estimation of $Y$. \\

In the TMLE-model, we try to identify all the sources of errors that will occur during this process impinging on the quality of all predicted values $\widehat{y}_{k}$ for the units $k\in U$ in our target population $U$ and their inclusion in the final statistics for the target population of interest.
%% My text lacked in undSerlining the importance of the fact that the TMLE still is about making 'predictions' in the final, finite, population. However, the text now oversees the fact that the parameters of the ML model are, in fact, estimates of parameters the infinite population.
% Action: bring in the tichotomy between these two stages and the fact that they talk about different populations
To accomplish this, the model is defined in two phases. The first phase, the training phase, tries to estimate the optimal set of parameters, $\widehat {{\boldsymbol{\Theta}}}$, based on the selected training set (see paragraph \ref{tr_ph}), whereas, during the application phase, the model is used to find an estimation of the target variable $\widehat Y$ (see figure \ref{fig:TMLE-appl} for an overview of the total model; paragraph \ref{appl_ph}). 
But first, we will focus on the estimation of the parameters $ \widehat { \boldsymbol {\Theta}} $ with its error term $\epsilon_{\widehat \Theta } $ .

\subsection{The training phase} \label{tr_ph}
\begin{figure}[htbp]
  \centering
  \includegraphics[width=1\textwidth]{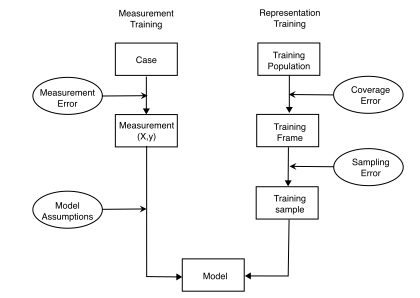} % Adjust the width as needed
  \caption{The Total Machine Learning Error model: training phase.}
  \label{fig:TMLE}
\end{figure}

The model also makes use of a measurement line and a representation line, as in the TSEM. The basic scheme for the training phase is depicted in figure \ref{fig:TMLE}.\\

\subsubsection{The measurement line}
The measurement line represents the evolution of variable values from its conception to its influence in the final model. It consists of two important steps within the training of an ML algorithm.

\paragraph{Case}
The first block at the upper left corner references each instance in the training set. The case still does not have any features, since these will be described in the next block (measurement). A case is best described as a pointer to an event or object in the real world that we have to make a prediction about. In the next step, we will observe and measure this object or event. 

\paragraph{Measurement}
In the process of measuring, the first errors will occur. We state here that these measurement errors are often neglected in data science, and can therefore be an important source of errors. We often hear many data scientists refer to the training set as the ground truth (true values in the original TSEM's terminology), and this is - by definition - wrong. In the case of supervised learning, for example, a training set has features and a target variable. Firstly, since the features ($\mathbf{X}$) can have measurement errors, they cannot be equal to their true values. We can write with little loss of generality $\mathbf{x} = \mathbf{x}^{(0)}+\boldsymbol{\epsilon}_{\mathbf{x}}$, where $\mathbf{x}^{(0)}$ denotes the true values (ground truth), $\mathbf{x}$ denotes the actually measured, distorted, values, and $\boldsymbol{\epsilon}_{\mathbf{x}}$ are the measurement error of variables $\mathbf{x}$.\\

Secondly, the model target variable $Y$ can also have errors.  We can also write $y = y^{(0)} + \epsilon_y$, where $y^{(0)}$ stands for the true value (ground truth), $y$ denotes the actually measured, distorted, values and $\epsilon_{y}$ stands for the measurement error of variable $y$. In a classification task, this leads to misclassifications (e.g.\ by a human annotator), which can be expressed using, e.g., one-hot encoding by $y\neq y^{(0)}$, where $y,y^{(0)}\in\{0,1\}^{\times n}$:\\
\begin{equation}
y^{(0)} = \begin{bmatrix}
    y^{(0)}_1 \\
    y^{(0)}_2 \\
    \vdots \\
    y^{(0)}_n
\end{bmatrix},
\end{equation}
where 
\begin{equation*}
\begin{cases}
    y^{(0)}_i = 1 & \text{if the case belongs to class $i$} \\
    y^{(0)}_i = 0 & \text{otherwise}.
\end{cases}
\end{equation*}

Let the transformation matrix $T$, where $T_{ij}$ describes the transformation probabilities from class $i$ to class $j$. The error in classifications can now be expressed in terms of the expected values of $y$ and this transformation matrix: 

\begin{equation}
\mathbb{E}(y) = \mathbb{E}(y^{(0)} + \epsilon_y) = T \cdot y^{(0)}
\end{equation}
This transformation matrix $T$ is, in fact, equal to a normalized confusion matrix.

\paragraph{Model (measurement perspective)}
Based on the preceding distorted data\-set $(\mathbf{x}, y)$, we have to form the model. At the moment that we have measurements (even with its measurement errors), we can start to train the model. Another source of errors we can identify is best described as the model assumptions. As is usual for models, the ML model is an abstraction of reality.  It will describe how certain variables are related to each other. This is based on the way the ML model assumes how these variables $(\mathbf{X}, Y)$ are related and how the features $\mathbf{X}$ are located in the feature space. Also, the choices made by the data scientist about which features he/she selects and how they are coded or transformed belong to the realm of model assumptions.\\ 

Hence, errors due to model assumptions are:
\begin{itemize}
\item errors due to the assumptions on how the features $\mathbf{X}$  are represented in the feature space (e.g. encodings);
\item errors due to the selection or actual availability of features $\mathbf{X}\neq\mathbf{X}^{(0)}$, where $\mathbf{X}^{(0)}$ stands for the true features generating the target variable $Y$ and $\mathbf{X}$ denotes the actual set of features used in the modelling exercise;
\item errors due to the assumptions in the functional dependence $f$, i.e. $f\neq\tilde{f}$, where $f$ stands for the true functional dependence and $\tilde{f}$ denotes the actually assumed dependence (to be estimated).
\end{itemize}

%% But this is a measurement error

All these assumptions degrade the model, but one should not forget that the idea of modelling is to generalize and the only way we can generalize is by making assumptions. The famous quote from George Box is highly relevant in this context. Box stated "All models are wrong, but some are useful. The question you need to ask is not 'Is the model true?' (it never is) but 'Is the model good enough for this particular application?'" \cite[p.~61]{Box2009}. In the end, it is all about the usefulness of the model, and a model with fewer erroneous assumptions is expected to be better. %% @all: is this true? it feels correct, but it should be argumented, PDAS: I hope that the quote and last added sentence helps.

After the model is defined and a functional dependence $\hat f$ is chosen, a parameter space is determined and errors, due to the optimization of model parameters ${\boldsymbol{\Theta}}$, i.e. $\widehat{{\boldsymbol{\Theta}}}\neq {\boldsymbol{\Theta}}^{(0)}$, where ${\boldsymbol{\Theta}}^{(0)}$ stands for the values for the parameters generating the smallest errors in target variable $Y$ from the features $\mathbf{X}$ and $\widehat{{\boldsymbol{\Theta}}}$ denotes the estimated (by optimization search, usually) values of these parameters. In many cases, a model is under-determined, leading to many equal optimal solutions ($\widehat{{\boldsymbol \Theta}}$ is in fact one instance from a set of solutions). This makes the solution not unique and could mean that, although the error is equal for different $\widehat{{\boldsymbol \Theta}}$, there are more stable points in the parameter space. A nice study about the fractal behavior of neural networks has been provided by \citet{sohldickstein2024boundary, sohldickstein20240212}.

\subsubsection{The representation line}
The representation line deals with errors regarding the concepts of population, frame, and sample (specific dataset for training, testing, and application). For ML models, these concepts are much subtler than in the TSEM, where the conceptions of set-theoretic population, frame, and sample are used (see section \ref{sec:DemMachine}). ML models are statistical models and the focus is on the joint distribution $F(Y, \mathbf{X})$ of the target variable and the features, which are usually decomposed as $F(Y, \mathbf{X})=F_{\boldsymbol{\Theta}}\left(Y|\mathbf{X}\right)F\left(\mathbf{X}\right)$ so that the focus is actually placed on the conditional dependence $F_{\boldsymbol{\Theta}}\left(Y|\mathbf{X}\right)$ \citep[see e.g.][]{HasTibFri09a}. In consequence, the conception of population, frame, and sample must be stated in terms of the generating distribution function, i.e.\ as an infinite population in the traditional jargon.\\

The dichotomy in these concepts lies much in line with the dichotomic distinction between enumerative and analytic surveys by \citet{Dem42a}. Indeed, in this work he introduced the concept of question A and question B types, noticing that each type has distinct implications for inference and decision-making. The production of official statistics engages mainly in type A questions, which involve actions based on existing data, intended to characterize and make decisions about known populations (set-theoretic conception). Conversely, question B types go beyond the limitations of current data, focusing on future measurements of unknown entities (statistical conception). To understand the inherent challenges of making inferences about unseen data, which is also an essential aspect of ML, we must first understand this dichotomy.\\

This dichotomy is not even novel in the current practice. Model-assisted estimation \citep[see e.g.][]{SarSweWre92a} makes an intelligent and profuse use of the dichotomy to introduce linear regression models (question B type) in the design-based inference paradigm (question A type).\\

Understanding the dynamics of populations is essential to understanding the difference between these two types of questions. Deming's 1941 paper, "On the Interpretation of Censuses as Samples" \citep{Deming41}, challenged the conventional view of a census as a complete enumeration of a population (set-theoretic definition). According to him, even a 100\% census represents a sample from a broader, infinite population. He argued that, if one needs to make an inference on a very small cohort of the complete population, even a 100\% sample could be too small to make that inference. Even though the sampling error will be zero, the more general type B questions cannot be answered based on this data. We need to consider the broader context when answering type B questions, i.e.\ we need to focus on the generating distribution function underlying the random phenomena behind the data generation.\\

In this line of thought, ML models and algorithms are generally designed to assist in solving problems similar to Deming's type B questions, aiming to predict (often future) instances of unobserved units based on observed data. As opposed to type A questions, which focus on actions based on existing information, type B questions involve anticipation and inference. Having understood this distinction, one can better understand the essence of ML, where predictive models are trained not simply to describe past events (although they can be used in this sense \citep[see e.g.][in this same book]{BarSaeSalSan24a}), but also to extrapolate patterns and relationships for informed decisions about data that has not yet been seen. As we shall argue, the problem of concept drift, data drift, or model drift originates here.\\

To illustrate our proposal let us focus on a concrete example where the dichotomy between set-theoretic and statistical concepts can be clearly observed. Let us try to construct a model-assisted estimator like the GREG estimator \citep[see e.g.][]{SarSweWre92a} but using a CART-type regression tree \citep[see e.g.][]{Mur13a} instead of a linear regression model\footnote{The argument remains valid for a linear regression model, but by using CARTs we hope to underline the different concepts involved, usually away from usual practice.}. Basically, we need to build a regression tree model (which involves both training and testing) to be applied to a concrete dataset to provide estimates for population totals of a continuous target variable $Y$. To be more specific, we may think of $Y$ as the turnover target variable in a periodic business statistics and the features $\mathbf{X}$ as the set of auxiliary variables available for the whole target population of interest (set-theoretic), as in the GREG scenario.\\

In terms of Deming's machine analogy, each box corresponds to actual data from a given reference time period. The machine corresponds to the country's economy producing these data for all periods. To build such a CART-assisted estimator amounts to investigating the statistical functional dependence $Y = f\left(\mathbf{X};{\boldsymbol{\Theta}}\right)+\epsilon$ behind the data generation mechanism $F_{\boldsymbol{\Theta}}\left(Y|\mathbf{X}\right)$ to provide estimates for, say, the population totals of the target population $U$ at a given reference time period $t$. As usual in the production of official statistics, we have a (set-theoretic) probabilistic sample $s_{t}$ selected according to a sampling design $p(\cdot)$ from the (set-theoretic) target frame $U_{Ft}$. For simplicity, we shall assume $U_{Ft}=U_{t}$ so there are no coverage errors affecting the target sample. We also consider there is no non-response. Errors from the measurement line are also considered non-present so that we can focus on the representation errors in the ML model. The CART-assisted estimator will be 

\begin{equation}\label{eqref:Cart_estim}
    \widehat{Y}_{dt}^{CART}=\sum_{k\in U_{dt}}\doublehat{y}_{k}+\sum_{k\in s_{d}}\frac{y_{k}-\doublehat{y}_{k}}{\pi_{k}},
\end{equation}

\noindent where $\pi_{k}$ stands for the first-order inclusion probability of unit $k$ and $\doublehat{y}_{k}$ is the design-based estimate of the CART-predicted value of variable $Y$ for unit $k$. The CART-predicted value $\widehat{y}_{k}$ can be written as \citep[see e.g.][]{Mur13a} $$\widehat{y}_{k}=\sum_{m=1}^{M}w_{m}I_{\widehat{R}_{m}}(\mathbf{x}_{k}),$$ where $\{\widehat{R}_{m}\}_{m=1}^{M}$ denotes the binary disjointly split regions of the feature space according to the estimated model and $w_{m}=\frac{1}{n_{m}}\sum_{k\in\widehat{R}_{m}}y_{k}$.\\

Now since we are taking a (set-theoretic) probabilistic sample $s_{t}$, as in the GREG estimation, we need to provide the design-based estimator so that

$$\doublehat{y}_{k}=\sum_{m=1}^{M}\left(\frac{\sum_{k\in\widehat{R}_{m}}y_{k}/\pi_{k}}{\sum_{k\in\widehat{R}_{m}}1/\pi_{k}}\right)I_{\widehat{R}_{m}}(\mathbf{x}_{k}).$$

We shall use this example to introduce the concepts of training and target populations, training and target frames, and training, test, and target samples.

\paragraph{Training Population}
% @David: i already wrote the subsections. Could you see if i interpreted the papers of Deming in the right way and if my cocnlusions are correct?
Let us denote by $F_{\boldsymbol{\Theta}}^{tr}\left(Y|\mathbf{X}\right)$ the data generation distribution function for those statistical units $U^{tr}$ used for training the model. This infinite population $F_{\boldsymbol{\Theta}}^{tr}\left(Y|\mathbf{X}\right)$ will be, in our TMLE-model, the training population. In our analogy using Deming's machine, the training population is a machine generating data later to be used just for training.\\

To assess its effect on the final statistics to be produced, we need to investigate its relationship with the final target population $U_{t}$ of interest. The difficulty arises because we need to compare this (set-theoretic) target population $U_{t}$ of interest with the (statistical) concept of training population $F_{\boldsymbol{\Theta}}^{tr}\left(Y|\mathbf{X}\right)$. We can consider $U_{t}$ as the realization of the underlying data generation distribution function $F_{\boldsymbol{\Theta}}\left(Y|\mathbf{X}\right)$ for the target population so that the finite population $U_{t}$ can indeed be conceived as a sample of this infinite target population (i.e.\ the superpopulation approach \citep[see e.g.][]{CasSarWre77a}).\\

In this line, the more different $F_{\boldsymbol{\Theta}}^{tr}\left(Y|\mathbf{X}\right)$ is from $F_{\boldsymbol{\Theta}}\left(Y|\mathbf{X}\right)$, the less accurate the final statistics will be. Although representativity as a comparison between two sets for estimating purposes constitutes a slippery concept \citep{KruMos79a,KruMos79b,KruMos79c,KruMos80a}, a mathematical definition in the set-theoretic realm \citep{Bet09a} can be provided in terms of the difference between (empirical) distribution functions for a given variable. In this same (to be made precise) we can talk of the representativity of the training population with respect to the target population in the context of an ML model in terms of the distance between $F_{\boldsymbol{\Theta}}^{tr}\left(Y|\mathbf{X}\right)$ and $F_{\boldsymbol{\Theta}}\left(Y|\mathbf{X}\right)$. The differences arise because of the changing dynamics of populations. For example, if data from the past are used, a judgement is implicitly made about the stability in time in the relationship between $Y$ and $\mathbf{X}$ for the target population at stake.

\paragraph{Training Frame}
Once the data generating distribution function $F_{\boldsymbol{\Theta}}^{tr}\left(Y|\mathbf{X}\right)$ is in place, data must be actually generated (Deming's machine must produce the bolts) so that we have a (set-theoretic) frame population $U^{tr}_{F}$ from which statistical units (instances) will be taken to train the model. This data-generating mechanism may be affected by different factors producing over-coverage, under-coverage, imbalance, etc. Errors can arise when the relationship between $Y$ and $\mathbf{X}$ is not extensively and properly covered throughout all generated instances. In mathematical terms, this means that it is impossible to obtain an accurate estimation of the training data generation distribution function $F_{\boldsymbol{\Theta}}^{tr}\left(Y|\mathbf{X}\right)$ from the frame population $U^{tr}_{F}$.

\paragraph{Target Sample}
The actual dataset used for training is composed of a selection of instances from $U^{tr}_{F}$. This selection may have been executed in many different ways, either producing a non-probability or (ideally) a probability sample $s_{tr}$.\\

Notice that even selecting a representative sample $s_{tr}$ with respect to the target frame $U^{tr}_{F}$, i.e.\ even having a small distance between the (empirical) distribution functions of the target variable in $U^{tr}_{F}$ and $s_{tr}$ this does not guarantee the final quality of the model if yet $F_{\boldsymbol{\Theta}}^{tr}\left(Y|\mathbf{X}\right)$ is very different to $F_{\boldsymbol{\Theta}}\left(Y|\mathbf{X}\right)$. However, to meet this final requirement, this notion of (set-theoretic) representativity of $s_{tr}$ with respect to $U^{tr}_{F}$ is necessary.

\paragraph{Model (representation perspective)}
In our previous sections, we discussed both the coverage error and the sampling error for the training phase. To create the optimal training set from the representation perspective to build the model we need to ensure that $F_{\boldsymbol{\Theta}}^{tr}\left(Y|\mathbf{X}\right)$ is very close to $F_{\boldsymbol{\Theta}}\left(Y|\mathbf{X}\right)$. The concept of representativity, as we have seen, is indeed slippery and involves diverse subtleties.\\

As a result of this procedure, we believe that problems like concept drift will be minimized to the best of our ability, as the final model will be robust to the dynamics occurring in the finite population under consideration.

\subsection{The testing phase} \label{tst_ph}
\begin{figure}[htbp]
  \centering
  \includegraphics[width=1\textwidth]{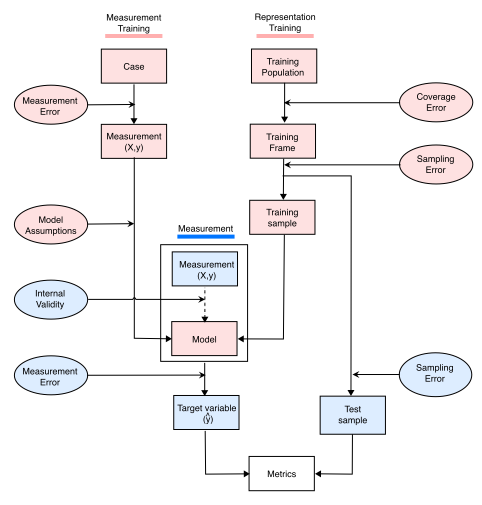} % Adjust the width as needed
  \caption{The Total Machine Learning Error model: training and testing phases. The training phase is highlighted in red; the testing phase is highlighted in blue.}
  \label{fig:TMLE-test}
\end{figure}

For the testing phase, the model also makes use of a measurement line and a representation line, as in the training phase. The basic scheme for the combination of the training and testing phases is depicted in figure \ref{fig:TMLE-test}.\\

A cautious reader may think that an infinite population $F^{test}_{\boldsymbol{\Theta}}(Y|\mathbf{X})$ may be needed for the testing phase so that both the measurement and representation lines can be defined in a similar fashion. However, this is in contrast to usual (good) practice. The available data are divided for training and testing, so that indeed we are ensuring  $F^{test}_{\boldsymbol{\Theta}}(Y|\mathbf{X})= F^{tr}_{\boldsymbol{\Theta}}(Y|\mathbf{X})$ (unless this division is executed in a highly non-ignorable way). This has a direct consequence for the scope of the validity of the tested models. Validity can be described as internal or external \cite{Onwuegbuzie2000}. The internal validity refers to the fact that the model learned the relationships that we were present in the training set, whereas the external validity refers to a validity beyond that point: the model is able to predict outside the context of the given population. This kind of validity is not reachable at the moment that $F^{test}_{\boldsymbol{\Theta}}(Y|\mathbf{X})= F^{tr}_{\boldsymbol{\Theta}}(Y|\mathbf{X})$. \\

It also has immediate consequences for both the measurement and the representation lines in the testing phase, as well as consequences for the scope of the validity of the models tested in this fashion.\\

\subsubsection{The measurement line}
The measurement line in the testing phase still represents the evolution of variable values since its generation to its influence in the final model, but now focused on those units used in the testing phase.

\paragraph{Measurement} 
When measuring variables for the testing phase, the same situation occurs as in the training phase: measurement of variables takes place with errors. Measured values are not in general equal to the corresponding true values.\\

Internal validity is indeed ensured when both training and testing data come from the same training target frame, as stated out above. This may not be the case, for example, when training data are taken from preceding time periods and testing data are used for the next period in the time series. The splitting into training and testing should be executed with extreme care to ensure that both underlying infinite populations are indeed the same!

\paragraph{Target variable} 
Once the model has been trained and tested, even with measurement errors, we can produce the predicted values $\widehat{y}$ since the functional dependence $\widehat{f}$ has been estimated as well as the parameters $\widehat{\boldsymbol{\Theta}}$.

\paragraph{Metrics (measurement perspective)}
Once predicted values can be produced, then model performance evaluation can be undertaken with the corresponding metrics. The basis for these kinds of validation measures lies in the well-known confusion matrix. It is important to underline that the model quality in its final application to the target population of interest will depend on the choice of metrics. The best reference to an overview of the confusion matrix and all its derived metrics can be found on Wikipedia  \citep{wiki:confusionmatrix}.

\subsubsection{The representation line}
Once the training/testing splitting is executed ensuring that both underlying infinite populations are the same, we can assume that the testing population and testing frame are the same as in the training phase so that only the samples will indeed be different: this is the unique novel element.

\paragraph{Test sample} 
The actual dataset used for testing is composed of a selection of instances from $U^{tr}_{F}$ according to the mentioned training/testing splitting strategy. This selection may have been also executed in many different ways, either producing a non-probability or an (ideally) probability sample $s_{test}$.\\

Under these assumptions, representativity properties of $s_{tr}$ and $s_{test}$ are shared, thus sampling errors are equally present in the testing phase, although it is common that the test set is much smaller than the training set. Particular training/testing splitting strategies may introduce differences between $s_{tr}$ and $s_{test}$. Cross-validation, out-of-bag procedures, and similar techniques are highly convenient and adequate in this sense.

\paragraph{Metrics (representation perspective)}
Model performance indicators and metrics are then computed for $s_{test}$. Notice that, when performance indicators and metrics on the test set are considered adequate, the whole model is retrained in the whole training/testing dataset, thus for the same underlying infinite population.

\subsection{The application phase} \label{appl_ph}
\begin{figure}[htbp]
  \centering
  \includegraphics[width=1\textwidth]{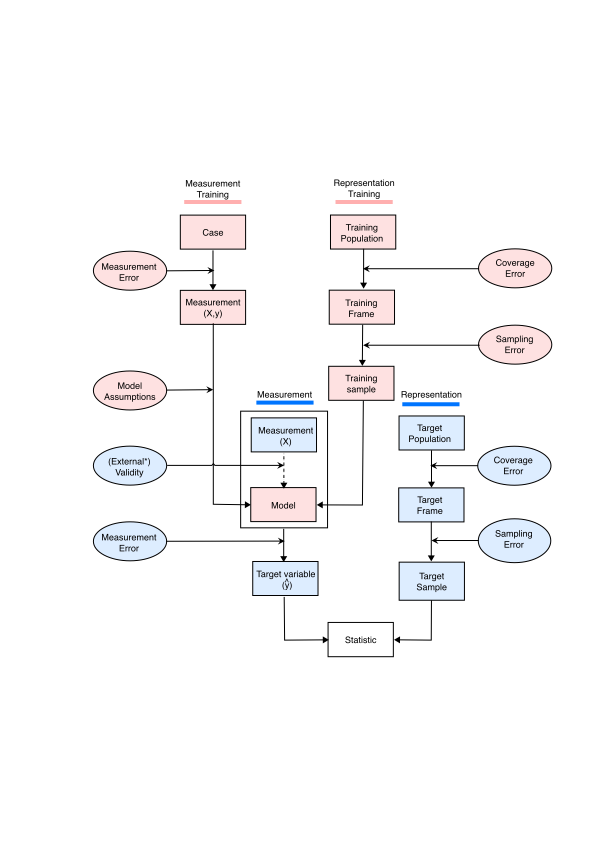} % Adjust the width as needed
  \caption{The Total Machine Learning Error model: training/testing and application phases. The training/testing phases are highlighted in red; the application/prediction phase is highlighted in blue.}
  \label{fig:TMLE-appl}
\end{figure}

Ultimately, one of the goals of creating an ML model is to use it on a set of new data. This is shown in the TMLE-model in figure \ref{fig:TMLE-appl}. The training phase (implicitly including the testing phase) is highlighted in red, whereas the prediction/application phase is highlighted in blue.\\

The central aspect of error assessment is that the model now includes two infinite populations, namely the training population $F^{tr}_{\boldsymbol{\Theta}}\left(Y|\mathbf{X}\right)$ and the infinite target population $F_{\boldsymbol{\Theta}}\left(Y|\mathbf{X}\right)$ and the realised finite target population $U$, which is the population for statistical analysis.\\

At this point, the original TSEM (figure \ref{fig:groves}) should be taken into account both for the variables $y$ (measurement line) and for the units $k\in U$ (representation line). As a final step, following equation \eqref{eqref:Cart_estim}, we will make an estimate based on collected target variable values $y$ and the predictions $\widehat{y}$ that have been made by the model.\\

It is important to note that we are now discussing the external validity of the model on the measurement side of the model. This is because we are applying the model to a population that is completely different from the one on which we trained and tested it. If the model does not perform as well as it did on the test set, it is not externally valid. As a form of validity, external validity is stronger than internal validity, and when applying ML algorithms, we should strive for an as good as possible external validity. In our opinion, this is also a better way to test the model. Instead of taking a sample of the same population as the training set, we should take a sample from a different (part of the) population. This would provide us with a much better assessment of the model.

\section{Summary of the model}
Having a comprehensive understanding of and meticulous rectification of inherent measurement errors prevalent in a variety of tasks is a critical aspect of ML endeavors. In addition to inaccuracies in feature measurement, these errors can also be attributed to the complexities associated with misclassifications of target variables, which can significantly impair ML accuracy and reliability. By meticulously acknowledging and adeptly mitigating these multifaceted errors, practitioners can substantially enhance the robustness and effectiveness of their models, resulting in greater trust and utility in their applications across various domains and scenarios.\\

A pivotal juncture occurs in the course of the Representation Dimension, as the focus shifts toward the intricacies of creating training populations, frames, and samples. Deming's seminal contributions to population analysis are underpinning this transition, whose distinction between type A and type B questions provides invaluable insight into population analysis's nuance. It is evident from Deming's distinction that it is paramount that training data accurately capture the subtle nuances and complexities inherent in the infinite population being studied. In addition, a growing emphasis is being placed on cultivating representative training frames that reflect the true essence of the broader population landscape as a resounding call to action among discussions regarding finite populations and their broad-ranging implications for model generalization.\\

Our focus is increasingly on rigorously evaluating model performance by applying a dedicated test dataset as we proceed through the testing phase. As practitioners navigate the labyrinthine complexities of measurement errors, they face a multitude of challenges, ranging from internal validity concerns to navigating the labyrinthine nature of internal validity concerns. In light of these challenges, robust evaluation methodologies that go beyond mere performance metrics, examining the intricate nuances of model behavior and efficiency across a variety of contexts and scenarios are imperative. Practitioners can increase their confidence and reliability in their ML endeavors by embracing the imperative of assessing external validity and ensuring that models can be applied to entirely new datasets.\\

In applying a model to previously unobserved data, we are presented with the ultimate frontier of ML deployment, wherein real-world applications and insights are derived from unobserved data, resulting in the fruits of labor. Iterative learning culminates in this phase, when models move from theory to practical application, enabling them to impact diverse domains and industries tangibly. There is a strong emphasis on external validity in this context, underlining the importance of models demonstrating robustness and generalizability across a variety of scenarios and contexts. In contrast to the confines of finite populations, Deming's principles promote models that transcend the confines of finite populations and embrace the dynamic complexity of the broader landscape by drawing parallels with the TSEM. For ML to reach greater heights of innovation and impact, practitioners must maintain a commitment to excellence and a steadfast commitment to advancing the frontiers of knowledge to navigate the rapidly evolving landscape with confidence and efficacy.\\

Based on these considerations, we would like to propose the following best practices:
\begin{itemize}
    \item Understand Measurement Errors: Thoroughly investigate and address measurement errors in feature variables and target variables during model training.
    \item Construct Representative Training Frames: Strive to create training frames that accurately represent the characteristics and complexities of the infinite population, considering diversity and coverage issues.
    \item Evaluate Model Performance: Use test sets to evaluate model performance, focusing on the external validity. Having a test set that is not part of the training population helps in determining the external validity
    \item Continuously Monitor and Refine Models: keep checking the validity of the training population that was used to create the model. Make sure that the target population is still a sample of the assumed infinite population. Iterate, when necessary on model development, incorporating feedback from evaluations on test sets and real-world applications to improve performance and robustness.
\end{itemize}

In the context of ML pipelines, we would like to emphasize the importance of integrating these insights into every stage of the development lifecycle. From data collection and preprocessing to model training and evaluation, incorporating robust mechanisms for addressing measurement errors and ensuring representativeness in training data is paramount. This entails implementing rigorous validation protocols, leveraging diverse datasets to capture the full spectrum of population characteristics, and continually refining models to enhance their generalizability and reliability.\\

Moreover, fostering a culture of transparency and accountability is essential, wherein practitioners actively document and communicate the limitations and assumptions underlying their models. By promoting open dialogue and collaboration, we can collectively identify and address potential biases and distortions, thereby fostering greater trust and confidence in ML applications.\\

Furthermore, investing in ongoing research and development efforts aimed at elucidating the intricate dynamics of population analysis and model evaluation is crucial. This entails exploring novel methodologies for assessing external validity, refining sampling strategies to minimize bias and variance, and advancing techniques for quantifying and mitigating measurement errors.\\

Ultimately, by integrating these best practices into ML pipelines, we can pave the way for the development of more robust, reliable, and ethical models that not only excel in performance but also uphold the highest standards of integrity and accountability. Through a steadfast commitment to excellence and a relentless pursuit of innovation, we can harness the transformative potential of ML to address some of the most pressing challenges facing society today, driving progress and prosperity for all.\\

\section{Applying Machine Learning models: some classification examples} \label{application}
As described above, creating a good externally valid ML model is essential when applying this kind of algorithm in the context of official statistics. Here, we will first describe the approach followed in the study performed to detect innovative companies, followed by the detection of online platforms. In the end, some recent insights gained during a detailed study of the creative industry are discussed. All studies use website text to identify different types of companies in the Netherlands. These studies were performed by some of the coauthors and nicely illustrate the insights gained during our study of ML-methodology.

\subsection{Detecting Innovative Companies}
Producing an overview of innovative companies in a country is a challenging task. Traditionally, this is done by sending a questionnaire to a sample of companies. This approach, however, focuses on large companies and completely misses small companies, such as startups. Therefore, an alternative approach was investigated by determining if a company is innovative by studying the text on its website. An ML model was developed based on the texts of the websites of companies included in the Community Innovation Survey of the Netherlands. The latter is a survey carried out every two years that focuses on the detection of innovative companies with 10 or more working persons. All websites of the innovative companies were included (a total of 3340) in addition to a similar sized-random sample of the non-innovative companies (3302) \citep{DaasDoef2020}. This provided the training frame which, according to the population topics discussed above, is very likely representative. It was found that the ML model developed was able to reproduce the results from the Community Innovation Survey, with a maximum accuracy of 93\%, and was also able to detect innovative companies with less than 10 employees, such as startups \cite{DaasDoef2020}. Manual checking was performed to determine the accuracy of the model on the classification of small companies and confirmed its external validity (regarding this subpopulation).\\

In a separate study, focused on misclassification errors, the model was applied to a very large, more representative, dataset of websites; of around 400.000. When a new model was trained on a random sample of 20.000 websites in the newly classified dataset, it became clear that even though the new model was trained on a more representative part of the (finite) population, the accuracy of the model had a decreased accuracy of 88\%. This demonstrated that misclassification errors build up during model development.\\

The downside of the original model was its stability over time. Here, a fairly rapid decline was observed on the same set of websites scraped at various points in time \cite{DaasJansen2020}. This is referred to as concept drift. Studies to reduce this issue were performed and found to be a challenging topic. Application of Large Language-based models provided the most successful 'solution' thus far with an accuracy of 86-87\% for the various data sets over time. However, the external validity of that model, on new unseen data, was not that high. It had an accuracy of 72\%.

\subsection{Detecting Online Platforms}
Obtaining reliable information from a small or rare subpopulation is a challenging topic. Approaches commonly used to find rare or so-called hard-to-identify groups are a screening survey, network sampling, area sampling, or a combination. An example of a rare subpopulation is online platform businesses. To get a complete overview of this subpopulation, an ML model was developed to identify these kind of platforms. This required a training frame. Hence, business statistics experts of Statistics Netherlands were asked to manually provide examples of such websites. This resulted in a set of 569 online platforms and 303 non-platform organizations, with very similar characteristics, that were additionally identified during this process. To the latter, a random sample of 266 non-platform organizations, from the websites linked to the Business Register, were additionally added.\\

The combined frame contained 50\% platform and 50\% non-platform websites used for model development. This resulted in an ML model that was able to identify online platforms with an accuracy of 82\% on the test set \citep{Daasetal2024}. After applying the model to the entire population of websites linked to the business register and manual inspection of random samples, it became clear that the model seriously overestimated the number of online platforms. In other words, the external validity of the model was low and its findings were biased. By sending questionnaires to a large sample of businesses this problem was initially solved and the answers obtained were used to validate the ML model-based findings \cite{Daasetal2024}.\\ 

The initial study revealed that online platforms merely compose 0.22\% of the total population of businesses with a website in the Netherlands. This made clear why false positives were such a huge problem. Subsequently, various steps were studied with the aim to considerably reduce the number of false positives detected by the ML model \citep{Gubetal2024}. The combination of steps that worked well are listed in Table \ref{tab:Table1}. \\

\begin{table}
    \centering
    \caption{Effect of various model-based approaches on online platform detection.}
    \begin{tabular}{lllcrcc}
    \hline
         &  &Type of model  &True Pos. &Est. Pos.  &Bias  &Accuracy \\ 
         \hline
         &1  &Log. reg.  &69  &2991  &0.098  &0.901 \\
         &2  &Log. reg. prob.  &69  &7657  &0.255  &0.901 \\
         &3  &Log. reg. cal. prob.  &69  &637  &0.019  &0.985\\
         &4  &Ensemble cal. prob.  &69  &306  &0.007  &0.993 \\
    \hline
    \end{tabular}    
    \label{tab:Table1}
\end{table}

Apart from producing binary labels, the model used could also produce the probability of a website being an online platform. These probabilities have a value between 0 and 1. Using the ‘probabilities’ of the model was found to increase the online platform estimates; hence, increasing the bias (Table \ref{tab:Table1}, row 2). This makes clear that, under these circumstances, the model did not produce actual probabilities. This did not affect the accuracy.\\ 

Subsequently, a method was applied to correctly calibrate the probabilities of the model used. This method corrects for the intrinsic prevalence of the model; i.e. the prevalence caused by the ratio of positives and negatives on which the ML model was originally trained (Puts and Daas, 2021). Since the number of online platforms is very low in the population and the model was trained on a much higher ratio of positive and negative cases (either 50-50\% or 30-70\%), it can be expected that correcting for this prevalence may seriously reduce the number of positive cases estimated. Applying the calibration method revealed that this was indeed the case; see Table \ref{tab:Table1}, row 3. As a consequence, the accuracy considerably increased.\\

When the results of multiple trained and calibrated models, up to 10, were combined, the bias was reduced even further; see Table \ref{tab:Table1}, row 4. The accuracy also increased somewhat. The bias is, however, not completely removed by the combination of correction methods applied. This is not unexpected for a model detecting rare events. We think there are two ways to even further improve this approach. The first is by increasing the number of models included in the ensemble. The second is by improving the representativeness of the websites included in the training frame used and those in the dataset used to produce the findings shown in Table \ref{tab:Table1}. This obviously relates to the discussion on infinite and finite populations mentioned above. 

\subsection{Detecting the Creative Industry}
Next, we discuss an ML study on the detection of businesses belonging to the creative industry in the Dutch municipality of Eindhoven. Since the creative industry is very difficult to define, it was interesting to study the topic with a data-driven approach. The big question in this study was - initially - if such businesses could be identified with an ML model trained in the texts on the websites of positive and negative cases. This required examples and, hence, local experts were asked to provide a list of websites of businesses belonging to the creative industry.\\

At the start of the study, a list of 110 positive websites was provided. Considering the number of positive cases included in the training frame in the cases discussed above, this is a very low amount. However, assuming that 'a website of a business belonging to the creative industry' is a rare event, the idea emerged that a (small) random sample of the websites linked to the business register, excluding the websites already in the positive set, could provide a nearly perfect list of non-creative industry examples. Such an approach is referred to as Positive and Unknown (PU)learning \citep{PUlearn} and might provide a solution. Subsequently, random samples were drawn from the websites of the business register (of various sizes), combined with the 110 positive cases and various models were trained. Be aware that the selection procedure used assured that no websites already included in the positive set were selected from the business register linked list of websites. After some trial and error, an ML model was found that seemed to produce fairly accurate results, on the test set, with an accuracy of 86\%. \\

Applying the model to the population, resulted in a probability distribution that was composed essentially of two clearly distinct peaks: a large one of non-creative websites (with an average of around 0.05) and a smaller group of potential creative websites (with an average probability of 0.99). However, manual inspection of a sample of 370 relatively high-scoring websites revealed that only 52\% of those websites actually belonged to the creative industry. The external validity of the model was obviously poor.\\

Since random samples were drawn for manual inspections, the idea emerged to add the findings for the websites to the training frame. Subsequently, in the next iteration, the combination of 110 positive and the 370 manually classified cases (including both negative and positive cases) was combined with various amounts of randomly sampled websites from the websites linked to the business register. Here again, the selection procedure used ensured that no websites already included in the positive and manually classified set were selected from the business register list of websites. This procedure resulted in a model with an accuracy of 85\% on the test set.\\

Applying this second model to the population revealed a probability distribution composed of three distinct groups: a very large group with an average probability of 0), a small group with an average probability of 0.1) and a small group with an average probability of 1). Manual inspecting random samples from each group revealed that each group contained, respectively, 2\%, 40\%, and 87\% websites belonging to the creative industry. Hence, the second model was much better able to discern creative industry websites compared to the first model. This leads to the conclusion that including the 370 manually inspected websites in the training frame, the 100 positive cases, and a random sample of unknown cases, made the resulting frame much more representative of the population studied. We think that following such an iterative approach is a very interesting way to create high-quality (and more representative) training frames with a fairly low manual effort. It reminds us of the Deming Cycle of continuous improvement. Because this study was performed during the development of the TMLE model described above, this also indicated that having such a frame in mind while performing an ML based study also helps to improve the quality of its findings.

\section{Discussion}
From the above, it is clear that the methodology of applying ML in official statistics is just in its infancy. Compared to the questions posed in \citet{Puts2021} this document already sheds some light on the methodology concerning the human annotation of data, sampling the population to obtain representative training sets, dealing with concept drift, and correcting the bias caused by the ML model. However, there are several additional and important considerations - from a methodological perspective - that became apparent while writing this document. These are at various stages of development and are described in the paragraphs below. Some of them are quite fundamental and all should be thoroughly investigated.\\ 

\textbf {(1) The terminology used by statisticians and data scientists differs.}
Historically, emerging fields use their own terminology. This is usually not problematic, since this terminology is typically only used within the newly defined paradigm. However, if the emerging field is adopted in another field, with its own paradigm, it will result in a loss of common ground. This does not necessarily have to result in a total misunderstanding between the fields. The diverging terminology within different fields has a striking similarity to the term "false friends"; which are words written exactly the same in two different languages but with a divergent meaning. Even though multilingual families have to deal with 'false friends' on a daily basis, they can function in peace and harmony without many misunderstandings. Why would this be a problem in science? The reason is evident: the older field (in this case statistics) will consider the terms used in the younger field (in this case ML) as incorrect. To deal with 'false friends', however, we should acknowledge their different meanings and keep them under consideration when communicating (like in bilingual families). Part of the purpose of this chapter was to acknowledge that methodology in statistics means something different compared to ML, and subsequently describe the field of ML in the terminology used within the field of official statistics.\\

\textbf {(2) Ensure homogeneity in the construct measured.} 
Developing an ML model starts with a training frame including cases of the best possible quality. To enable this, there usually is a "human in the loop", for instance, to ensure that the cases included are correctly classified or to check the findings of the model on new data. Including human checking is challenging because the findings of multiple humans need to be consistent, the so-called inter-annotator agreement, which requires considerable effort. Efficient ways to verify the construct measured by ML models need to be developed.\\  

\textbf {(3) Representativity in the context of ML.}
This fundamental and challenging question touches the heart of statistical inference. There is a definitive need to develop more theory in this area to ensure that ML models, as accurately as possible, measure the concept of interest on new cases. Which steps should be taken, in which order, to enable this? In many of the questions included in this list, the notion of representativity in the context of ML is paramount and should therefore be resolved. \\

\textbf {(4) How to deal with ML and type A and type B questions?} 
Deming distinguishes these types of questions. Type A questions are action-based on existing information and more in line with the production of official statistics, while type B questions aim to predict instances of unobserved units and go beyond the limitations of current data. Here, one could simply decide to just focus on ML and type A questions, but one needs to be aware that ML models developed for answering type B questions are expected (when properly developed) to better deal with new data and changing conditions. Both types are highly relevant and ML models that can deal with both should be studied.\\ 

\textbf {(5) Bias(es) resulting from errors made during training of the model (especially for type B questions).}  
Many of the errors occurring during model development will result in biased estimates. Reducing the errors as much as possible during model development is a way to decrease their disruptive effect. The TMLE-model is a good starting point here and the focus should now shift to measure each of the errors identified. Of course, one should bear in mind that this is not as evident as it seems. For instance, errors introduced by model assumptions are almost impossible to quantify.  \\   

\textbf {(6) Representativity problems resulting from differences in the population composition of the training frame and the infinite population.} 
We saw that machine learning involves answering a type B question (see previous point). Thus, the training set should not be representative with respect to the finite population, but rather with respect to the infinite population. 
Procedures are needed to ensure that the cases in the training frame include the relevant features occurring in the infinite population as well as possible. It's obvious that random sampling will be an important step here. However, what is the best approach to obtain such a training frame? certainly for topics focused on rare events.\\

\textbf {(7) How to approximate the infinite population?} 
A possible approximation of the infinite population can be done by taking one simple random sample: the finite population. We will, however, introduce parts in the feature space that are not well represented in this final population. The ill-covered finite population (with respect to the infinite population) can be seen as a coverage error, but also as a sampling error. Acknowledging this error is already an important step. However, the question remains how we could approximate the infinite population? To answer this question, we need to have a better understanding of representativity in the context of machine learning. Deming suggested approximating the infinite population by taking several finite populations, separated in time to ensure independence between the 'samples', but this approach is not always feasible. Consequently, the question remains and needs to be answered.\\

\textbf {(8) Should we sample with replacement or not?} 
When reading papers of Deming on the subject of infinite populations, it becomes clear that, in order to minimize the errors on the estimators (in our case: the parameters of the model), we need to sample with replacement. This is not an approach commonly applied in ML. Here, the training set is sampled from the training frame without replacement in order to minimize the overlap between elements in the feature space. Its clear that there is a problem here. Future work is needed to better understand this fundamental question. \\ %eigenlijk wil je hier geen overlap/duplicaten, dat zorgt namelijk voor te hoge accuracies 
% ik denk dat daar de vraag over gaat, inderdaad. 

\textbf {(9) Develop a procedure that ensures that only the most important features (variables) are included in the model.}
When larger training frames are being used in ML model development, we have observed that increasing numbers of features become included in models. There is a definite need for an approach that reduces this effect and ensures that only the 'best' features are selected. Such a procedure could also improve the accuracy and stability of the model over time.\\ 

\textbf {(10) Develop a procedure that ensures that the ML model is both internally and externally valid and as stable as possible over time.}
From the above, it has become obvious that the ultimate goal when developing an ML model is creating a model with a high external validity. This is linked to topic (6). There is a definitive need for a procedure that ensures a model is obtained with both a high internal and external validity. The findings of topic (6) will certainly help here.\\    

These topics go beyond the focus of many traditional ML practitioners and highlight the importance of a statistical view on ML and the need for ML methodology. We are convinced that the TMLE model, in particular the sources or errors identified, will help users to get a better grip on the challenging application of ML in a statistical context but also when applying ML in general. In addition, the document gives an overview of the topics that need to be studied in more detail and, as such, sets the stage for future research in this interesting and challenging area.\\

\section*{Acknowledgements}
First of all, we would like to extend our heartfelt thanks to Yvonne Gootzen for her invaluable contributions to this work. We are deeply grateful for her involvement.

We also wish to acknowledge Luuk Gubbels and Sanne Peereboom, whose internship work provided valuable insights that contributed to section \ref{application}. Delorian Canlon, and Sourav Bhattacharjee are gratefully acknowledged for stimulating discussions that considerably improved the document.

\bibliographystyle{chicago}
\bibliography{main.bib}

\end{document}